\PassOptionsToPackage{table}{xcolor}
\documentclass[manuscript]{acmart}
\usepackage{graphicx}
\usepackage{amsmath}

\usepackage{amssymb}
\usepackage{pifont}
\newcommand{\cmark}{\color{green}\ding{51}}%
\newcommand{\xmark}{\color{red}\ding{55}}%
\usepackage{booktabs}
\usepackage{microtype}
\usepackage{setspace}
\usepackage{multicol}
\usepackage{multirow}
\usepackage{xspace}
\usepackage{hyperref}
\usepackage{listings}
\lstdefinelanguage{json}{
  morestring=[b]",
  showstringspaces=false,
  breaklines=true,
  morecomment=[l]{//},
  literate=
   *{0}{{0}}1
    {1}{{1}}1
    {2}{{2}}1
    {3}{{3}}1
    {4}{{4}}1
    {5}{{5}}1
    {6}{{6}}1
    {7}{{7}}1
    {8}{{8}}1
    {9}{{9}}1
    {:}{{:}}1
    {,}{{,}}1
}
\usepackage[T1]{fontenc}
\usepackage[most]{tcolorbox}
\definecolor{PromptBlue}{RGB}{28,107,160}
\definecolor{PromptBG}{RGB}{239,246,252}
\newtcolorbox{llmPromptBox}[2][]{%
  enhanced,
  breakable,
  colback=PromptBG,
  colframe=PromptBlue!60,
  coltitle=black,
  title={#2},
  fonttitle=\bfseries,
  arc=2mm,
  boxsep=2.5mm,
  left=3mm,right=3mm,top=2.5mm,bottom=2.5mm,
  attach boxed title to top left={xshift=3mm,yshift*=-3mm},
  boxed title style={
    colback=PromptBlue!15,
    colframe=PromptBlue!60,
    boxrule=0.5pt,
    arc=2mm,
    top=1mm,bottom=1mm,left=2mm,right=2mm
  },
  #1
}
\usepackage[acronym]{glossaries}
\glsdisablehyper

\makeglossaries
\newacronym{llms}{LLMs}{Large Language Models}
\newacronym{ai}{AI}{Artificial Intelligence}
\newacronym{idc}{IDC}{International Data Corporation}
\newacronym{kear}{KEAR}{Knowledge Elicitation and Retrieval}
\newacronym{clsd}{CLSD}{Cross-Lingual Stance Detection}
\newacronym{hci}{HCI}{Human-Computer Interaction}
\newacronym{cli}{CLI}{Command Line Interface}
\AtBeginDocument{%
  }

\setcopyright{acmlicensed}
\copyrightyear{2018}
\acmYear{2018}
\acmDOI{XXXXXXX.XXXXXXX}
\acmConference[Conference acronym 'XX]{Make sure to enter the correct
  conference title from your rights confirmation email}{June 03--05,
  2018}{Woodstock, NY}
\acmISBN{978-1-4503-XXXX-X/2018/06}

\begin{document}

\title{Can AI Guess What You Know? Performance Comparison of Large Language Models for Human Domain Knowledge Estimation From Communication Logs}

\author{Ko Watanabe}
\orcid{0000-0003-0252-1785}
\affiliation{%
  \institution{DFKI}
  \city{Kaiserslautern}
  \country{Germany}
}
\email{ko.watanabe@dfki.de}
\authornotemark[1]

\author{Shoya Ishimaru}
\orcid{0000-0002-5374-1510}
\affiliation{%
  \institution{Osaka Metropolitan University}
  \city{Osaka}
  \country{Japan}
} 
\email{ishimaru@omu.ac.jp}

\renewcommand{\shortauthors}{Watanabe et al.}

\begin{abstract}
Employees often struggle to identify ``who knows what,'' leading to organizational productivity losses. We investigate whether Large Language Models (LLMs) can infer individual domain knowledge directly from long-term Slack logs. Analyzing 27,188 messages from 43 users, we evaluated seven models (including Gemini, Claude, and GPT families) by comparing their zero-shot estimates against self-reported skill ratings from 27 participants. Gemini 2.5 Flash achieved the lowest error (MAE 21.13\%), while GPT models showed significantly larger discrepancies. Notably, estimation accuracy depended only weakly on message volume, indicating that more text alone does not guarantee better inference. These findings demonstrate the feasibility and current limits of automated expertise mapping, highlighting the need for privacy-preserving deployments and richer, structure-aware representations of human knowledge.
\end{abstract}



\begin{teaserfigure}
  \centering
  \includegraphics[width=\textwidth]{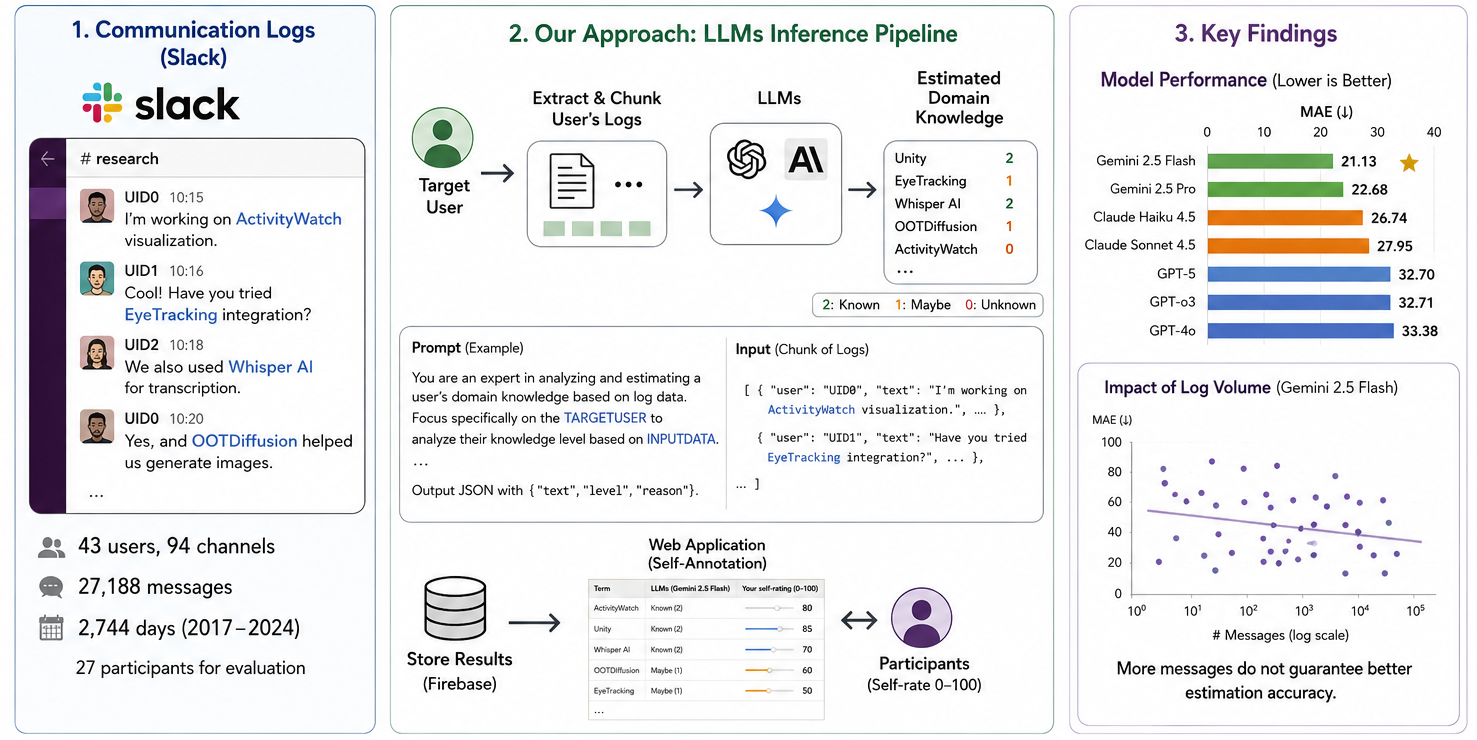}
  \caption{Concept image of this study. This study aims to estimate the domain knowledge of a person from their communication logs with \gls{llms}. In this study, we use the Slack communication logs as input and to estimate and visualize human domain knowledge using \gls{llms}.}
  \Description{Concept image of this study.}
  \label{fig:teaser}
\end{teaserfigure}


\maketitle

\section{Introduction}
Imagine you are a new worker in a company.
You join a group full of experienced workers.
Some know specific software technologies well, and some may know well about the credentials for a project.
Since you are new, you are not sure who the best person to ask is.
In the end, the approach may be to ask various people until you find the best solution.
This may also occur for a new student at a university or in a laboratory.
The issue does not sound like a huge problem, but it actually is, according to the various social costs.

Internal knowledge, so-called intranet sharing in a company, has a high failure rate, which \gls{idc} Fortune reported in 2017 in their white paper: 500 companies lose at least 31.5 billion USD a year~\cite{zohuri2019model, trippe2022knowledge, west2018future}.
McKinsey also reported that 1.8 hours (19\%) of the total work hours are spent searching for information or a person to ask for help~\cite{bughin2012capturing}.
Especially in the onboarding phase, \citet{PanoptoYouGov2018} reported that new employees can fully catch up and work independently within 6 months, at an estimated cost of 253 thousand USD.
As these reports suggest, a hidden intranet incurs high economic costs across organizations worldwide.

\gls{llms} leverage their strengths in processing and generating complex language, rapidly becoming versatile tools across various fields~\cite{rekimoto2025gazellm, zhou2025augmented, salminen2025use, oomori2024skillsinterpreter, suzawa2025augmenting}.
In fields such as healthcare~\cite{liu2025stopgap, takita2025comparative} and education~\cite{morita2025genair, chen2024bidtrainer, yamaoka2025img2vocab, yamaoka2023experience}, they contribute to organizing and interpreting vast amounts of information~\cite{yang2024harnessing}, from analyzing clinical records and massive medical datasets to generating human, understandable summaries and responses with performance often approaching human levels~\cite{mumtaz2023llms, clusmann2023future}.
Within organizations, interest is growing in utilizing \gls{llms} to quantify and share individual expertise~\cite{zhang2024imperative, freire2023tacit, wu2025llms}.
\gls{llms} are also being used to transform the content of communications by replacing or refining human utterances~\cite{gu2021dialogbert, galimzhanova2023rewriting, zhang2022analyzing}, such as by summarizing long multi-participant chat threads into concise highlights~\cite{kosilova2024survey} and rephrasing a user's statements in real time better to fit the audience or context~\cite{kumar2025leveraging}. 
These emerging applications highlight how \gls{llms} can measure human knowledge by extracting and reformatting information for improved understanding.

In this study, we investigate whether humans' individual domain knowledge can be estimated from chat logs with \gls{llms}. 
Our study uses Slack~\footnote{\url{https://slack.com/}} organizational communication logs as input and to estimate and visualize human domain knowledge using \gls{llms}.
To evaluate performance, we conduct a user self-annotation task after \gls{llms} estimation, asking users to rate the level of understanding of the domain knowledge extracted by the system.
By analyzing the gap between the estimated domain knowledge of various \gls{llms} and the user's self-annotated domain knowledge, we verify how well \gls{ai} perform in domain knowledge estimation using chat/communication logs.
Our key research questions (RQs) are as follows:

\begin{itemize} 
  \item[RQ1] How precise can \gls{llms} estimate the human domain knowledge?
  \item[RQ2] Which \gls{llms} model provide the most accurate knowledge estimation?
  \item[RQ3] How does the amount of communication logs impact the accuracy of \gls{llms} domain knowledge estimation?
\end{itemize}

This study will contribute to the field of semi-automated human domain knowledge estimation ecosystem. 
As we conceive, the organization can manage knowledge through daily activities, such as chat communication, to map team members' domain knowledge.

\begin{table}[t!]
  \centering
  \renewcommand{\arraystretch}{1.0}
  \caption{
    Comparison of related works against our proposed work. Our work is most closely related to \citet{zhang2024imperative}; however, the study is a survey paper that has not conducted any practical data analysis. Hence, our study is the first to estimate domain knowledge via chat-based LLMs.
  }
  \label{tb:related_work_comparison}
    \begin{tabular}{lcccccc}
      \toprule
        \textbf{Related Work}          & Survey & AI/ML  & LLMs   & Domain Knowledge Extraction & Chat   & Source Code \\
        \midrule
        \citet{kosilova2024survey}     & \cmark & \xmark & \xmark & \cmark                      & \cmark & \xmark \\
        \citet{zhang2024imperative}    & \cmark & \cmark & \cmark & \cmark                      & \cmark & \xmark \\
        \citet{huang2007extracting}    & \xmark & \cmark & \xmark & \cmark                      & \cmark & \xmark \\
        \citet{wang2024towards}        & \xmark & \cmark & \xmark & \cmark                      & \xmark & \xmark \\
        \citet{zhang2024llm}           & \xmark & \cmark & \cmark & \cmark                      & \xmark & \xmark \\
        \citet{tigunova2020extracting} & \xmark & \cmark & \xmark & \cmark                      & \cmark & \xmark \\
        \citet{arsovski2019automatic}  & \xmark & \cmark & \xmark & \cmark                      & \cmark & \xmark \\
        \midrule
        \textbf{Our Work}              & \xmark & \cmark & \cmark & \cmark                      & \cmark & \cmark \\
        \bottomrule
      \end{tabular}
\end{table}

\section{Related Work}
\autoref{tb:related_work_comparison} shows the related work and position of this study.
This section will explain what kind of related work exists in this field and the originality of our work.

\subsection{Survey on Organizational Chat Conversation Analysis}
\citet{kosilova2024survey} conducted a large-scale survey of organizational chat conversation analysis.
The study is a survey paper, so the actual dataset is not used for the practical case.
The survey selected 16 papers, and the conclusion stated that the domain significantly impacts the performance of knowledge elicitation, particularly in medicine and software development, which are often difficult.
\citet{zhang2024imperative} presents a comprehensive survey of conversation analysis in the era of \gls{llms}, formalizing it as a four-stage process encompassing scene reconstruction, causality analysis, skill enhancement, and conversation generation. 
Their work highlights the field's fragmentation, noting that existing studies predominantly address shallow subtasks such as emotion or intent classification while lacking deeper reasoning about conversational dynamics. 
They further identify the need for benchmarks and methods that capture goal-directed, multi-turn conversational behavior, underscoring substantial gaps between current research and real-world applications.

The communication logs for knowledge elicitation have been discussed in a survey paper, however, none of the studies actually use them in a practical case.

\subsection{Practical Case of Knowledge Elicitation from Communication Logs}
\citet{huang2007extracting} proposes a cascaded framework for automatically extracting high-quality <thread-title, reply> pairs from online discussion forums as chatbot knowledge. 
By combining SVM-based relevant-reply identification with ranking SVMs to select informative, concise, and trustworthy responses, the method effectively filters noisy forum content and surfaces reusable conversational knowledge. 
Experiments on a large movie forum demonstrate that the approach yields high-precision chatbot response pairs, substantially outperforming baseline methods.

\citet{wang2024towards} has mentioned the concept of Human-AI mutual learning, where \gls{ai} and humans learn from each other.
The unique point of this study is that explainable \gls{ai} is used to provide transparency into how \gls{ai} acquires new knowledge and to return the knowledge elicitation flow to humans.
Since the paper is a position paper, the actual dataset is not used for the practical case.

\citet{zhang2024llm} proposes \gls{kear}, an \gls{llms}-enabled knowledge elicitation and retrieval framework for zero-shot cross-lingual stance detection, addressing the challenge of transferring stance-relevant reasoning across languages with no target-language training data. 
The method elicits background, inference, and explanation knowledge from \gls{llms} reasoning, verifies them via multi-agent collaboration, and retrieves the most relevant knowledge through a hierarchical cross-lingual retriever. 
Experiments on multilingual benchmarks show that \gls{kear} significantly outperforms competitive zero-shot and even supervised \gls{clsd} methods, demonstrating the effectiveness of \gls{llms}-derived inferential knowledge for bridging language gaps.

\citet{arsovski2019automatic} presents a methodology for automatically extracting conversational knowledge from existing rule-based chatbots by sending large-scale question inputs and identifying the stable set of unique response rules they contain. 
The authors demonstrate that chatbot knowledge converges after sufficient probing and further validate this saturation point through K-means clustering over extracted responses. 
Using the obtained knowledge, they train a seq2seq neural conversational agent that reproduces the original chatbot's behavior, achieving high BLEU similarity and demonstrating effective machine-to-machine knowledge transfer.

We have discovered that a combination of \gls{llms} and chat (Slack) logs for knowledge extraction does not yet exist, and that is our primary focus of the study.


\begin{figure*}[t!]
  \centering
  \includegraphics[width=\textwidth]{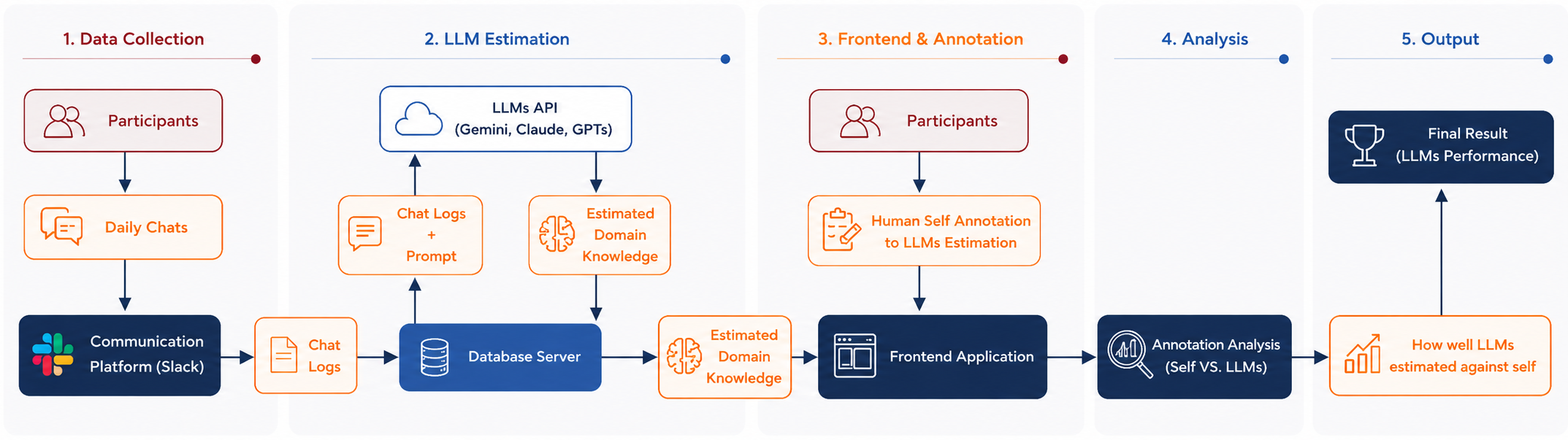}
  \caption{
    Overview of the proposed workflow.
    The first step is to export Slack communication logs as JSON files. 
    Then, the backend server reads the JSON files and uses them to generate a prompt when making a prompt request to \gls{llms}. 
    The \gls{llms} will generate an estimate of domain knowledge based on the system prompt and the user's message.
    Estimated domain knowledge is then stored in the Firebase Cloud Database~\cite{firebase_2025}.
    In the frontend web application, look at the cloud database and display each user's domain knowledge.
    When a user logs in to the web application, they can see their own domain knowledge.
    In the end, the user can also make their own annotations for each estimated domain of knowledge.
  }
  \Description{
    Overview of the workflow of the proposed method.
    The first step is to export Slack communication logs as JSON files. 
    Then, the backend server reads the JSON files and uses them to generate a prompt when making a prompt request to \gls{llms}. 
    The \gls{llms} will generate an estimate of domain knowledge based on the system prompt and the user's message.
    Estimated domain knowledge is then stored in the (Firebase) cloud database.
    In the frontend web application, look at the cloud database and display each user's domain knowledge.
    When a user logs in to the web application, they can see their own domain knowledge.
    In the end, the user can also make their own annotations for each estimated domain of knowledge.
  }
  \label{fig:architecture}
\end{figure*}

\section{Methodology}
\autoref{fig:architecture} shows the overview of the proposed workflow.
In this section, we will explain each component of the system architecture in detail.

\subsection{Statistics of Communication Log Dataset}
\label{subsec:log_statics}
In this study, we use the Slack communication logs.
Data was collected from April 30th, 2017, to November 4th, 2024 (2744 days).
The dataset contains communication logs from a company, totaling 27,188 messages.
There were 43 users in the chats and 94 channels.
We will explain the process of selecting specific data from this raw data and provide a detailed introduction to the data format.

\begin{figure*}[t!]
  \centering
  \includegraphics[width=0.9\textwidth]{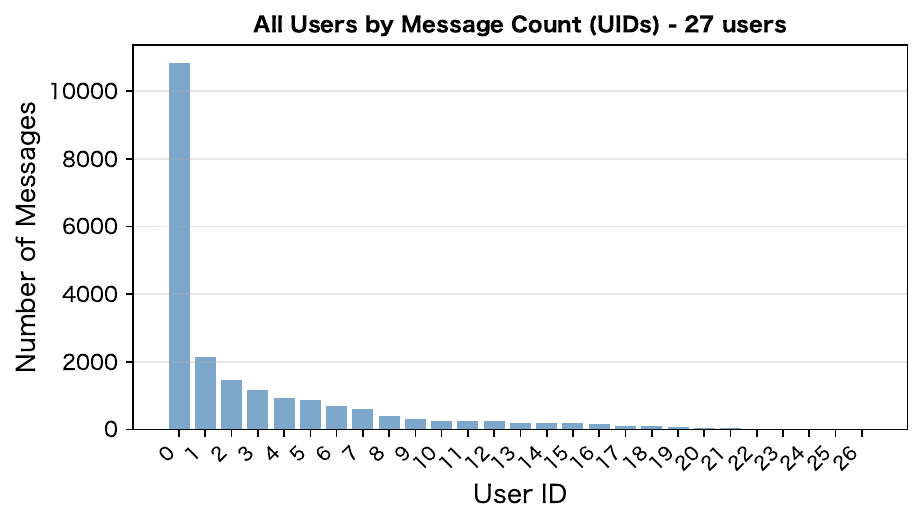}
  \caption{
    Number of messages per user (UID). 
    The quantity varies among users, so we examine the impact of this variation on estimation performance.
    The largest number of messages is generated by UID 0, and the smallest by UID 26.
    }
  \Description{
    Number of messages per user (UID). 
    The quantity varies among users, and so we check the impacts of this variation on the estimation performance.
    The largest number of messages is done by UID 0, and the smallest is done by UID 26.
    }
  \label{fig:number_of_messages}
\end{figure*}

\subsubsection{Selected Data Volume and Statistics}
Among 43 users, we selected 27 as the actual target participants for the study.
This is mainly due to the contact's availability, meaning some candidates have not been able to reach them.
\autoref{fig:number_of_messages} shows the number of messages per user (UID).
The largest number of messages is 10,819, done by UID 0, and the smallest is 3, done by UID 26.
The mean message count is 792, and the median is 208.
While user message volume varies significantly, the logs capture a broad range of user information to compare message volume and inference accuracy, the focus of this research.

\subsubsection{Data Structure}
\autoref{lst:slack-json} shows the structure of the Slack message entry.
The data is stored in a JSON format.
The key message is stored in the ``text'' field, and users who react to the message are also visible.
\autoref{lst:channel-join} shows the structure of a Slack channel join event message.
This message is not a user message, but a system message.
The user who joined the channel is visible in the ``user'' field.
We use this message to determine whether the user is in the channel.
This is an important point because we wanted to infer domain knowledge not only from the user's own utterances, but also from passive information the user observed during the conversation.

\begin{figure*}[t!]
  \lstset{basicstyle=\ttfamily\small, breaklines=true}
  \begin{lstlisting}[language=json, caption={Raw data of a Slack message entry. Some parts are anonymized.}, label={lst:slack-json}]
{
    "user": "UID0",
    "type": "message",
    "ts": "1683702597.263009",
    "client_msg_id": "...",
    "text": "... A hint for surveys: If you want to search related work around our research field, please search  \"keywords + conference name\". CHI, ETRA, UbiComp, ISWC, UIST, CVPR, AHs, SIGGRAPH, ...",
    "user_profile": {...},
    "thread_ts": "1683702597.263009",
    "reply_count": 2,
    "replies": [{"user": "UID1","ts": "1683704080.511989"}, ...],
    "reactions": [{"name": "+1","users": ["UID2","UID3"],"count": 2}],
    "attachments": [{"from_url": "...", "message_blocks": [...]}],
    "blocks": [{"type": "rich_text", ...}]
}
  \end{lstlisting}
\end{figure*}

\begin{figure*}[t!]
  \lstset{basicstyle=\ttfamily\small, breaklines=true}
  \begin{lstlisting}[language=json, caption={Raw data of a Slack channel join event message.}, label={lst:channel-join}]
{
    "subtype": "channel_join",
    "user": "UID4",
    "text": "<@UID4> has joined the channel",
    "type": "message",
    "ts": "1493555632.223680"
}
  \end{lstlisting}
\end{figure*}

\subsection{Selection of Large Language Models}
In this study, we selected the following \gls{llms} for the evaluation: Claude Haiku 4.5~\cite{anthropic_claude_haiku4_5}, Claude Sonnet 4.5~\cite{anthropic_claude_sonnet4_5}, Gemini 2.5 Flash~\cite{deepmind_gemini_2_5_flash}, Gemini 2.5 Pro~\cite{deepmind_gemini_2_5_pro}, GPT 4o~\cite{openai_gpt_4o_2024}, GPT o3~\cite{openai_o3_o4mini_2025}, and GPT 5~\cite{openai_gpt5_2025}.
We select the models that are API-friendly as of October 2025.

\paragraph{Claude Haiku 4.5:} 
The model is Anthropic's small, fast hybrid-reasoning large language model, designed to deliver near-frontier performance for coding, tool use, and computer control with much higher cost- and latency-efficiency than larger Claude models.
It is trained on a filtered mixture of public web data up to February 2025, licensed and partner datasets, opt-in user data, and synthetic data. It supports extended-thinking mode and a 200k-token context window for complex, multi-agent workflows.
Extensive internal and third-party evaluations indicate substantially improved alignment and robustness over Claude Haiku 3.5, strong safeguards against misuse in agentic scenarios, and CBRN capabilities below AI Safety Level-3 thresholds, leading to deployment under Anthropic's ASL-2 standard.

\paragraph{Claude Sonnet 4.5:}
The model is Anthropic's latest hybrid-reasoning large language model, with an extended-thinking mode and state-of-the-art performance on software engineering, long-horizon agentic workflows, and real-world computer-use tasks, while also improving general reasoning and mathematics.
Extensive pre-deployment evaluations covering safeguards, agentic safety, cybersecurity, reward hacking, alignment, and model welfare show substantially improved safety and honesty leading to deployment under Anthropic's \gls{ai} Safety Level 3 standard.
Taken together, these results position Claude Sonnet 4.5 as Anthropic's primary high-intelligence model, combining cutting-edge coding and agent capabilities with conservative, policy-driven safety controls suitable for safety-critical and scientific applications.

\paragraph{Gemini 2.5 Flash:}
The model is a next-generation lightweight Gemini model that lets users control its ``thinking budget'' to trade off reasoning depth against latency and cost for high-throughput applications. 
It is natively multimodal, jointly processing text, audio, images, and video, with a 1-million-token context window that enables exploration of huge datasets and long-horizon interactions in a single session. 
The model also supports native audio outputs and seamless switching among 24 languages with the same voice, making it suitable for expressive, interactive, and globally deployed AI systems.

\paragraph{Gemini 2.5 Pro:}
The model is a reasoning model, designed to solve complex problems and to understand vast datasets spanning text, code, audio, images, video, and even entire code repositories.
It is a multimodal model deployed on Vertex AI with a 1,048,576-token input window and 65,535-token output capacity, supporting a wide range of enterprise and research workloads.
The model exposes advanced capabilities, such as tool use (e.g., code execution and RAG with the Vertex AI RAG engine), system instructions, function calling, and structured outputs, making it suitable as a general-purpose backbone for sophisticated AI agents. 

\paragraph{GPT 4o:}
The model is an end-to-end multimodal model capable of processing and generating text, audio, and images within a single unified neural architecture.
It delivers human-like response latency in speech interactions and achieves GPT-4-level performance while improving speed, multilingual ability, and vision/audio understanding.
The model incorporates extensive safety evaluations and mitigations, including red-teaming, content filtering, and controls for unauthorized voice generation.

\paragraph{GPT o3:}
The model is a reasoning-focused models that combine advanced chain-of-thought reinforcement learning with full tool capabilities, enabling strong performance in math, coding, scientific analysis, and multimodal tasks.
The models enhance safety through deliberative alignment and an instruction hierarchy that prioritizes system-level constraints.
Extensive evaluations show improved robustness against harmful content, jailbreaks, and ungrounded inferences, while maintaining state-of-the-art reasoning and tool-use efficiency.

\paragraph{GPT 5:}
The model is a unified AI system that combines a fast, high-throughput model, a deeper GPT-5 thinking reasoning model, and a real-time router that chooses between them based on task complexity, tool needs, and user intent.
It delivers state-of-the-art performance across coding, math, writing, health, and visual perception, while reducing hallucinations and improving instruction-following and overall usefulness in real-world ChatGPT queries.
GPT-5 further introduces output-centric ``safe-completions'' training, extensive red-teaming, and biological/cybersecurity preparedness safeguards to limit harmful use, decrease sycophancy and deception, and better handle dual-use and safety-critical scenarios.

\subsection{Domain Knowledge Extraction Workflow}
In this study, we provide a \gls{cli} pipeline that extracts user domain knowledge from Slack archives.
It starts by loading environment variables for three \gls{llms} providers (OpenAI, Claude, Gemini) and by exposing helpers that ingest Slack member metadata, allowing filtering by billing status or activity. 

The \gls{cli} relies on argparse to capture user IDs, filter settings, dataset roots, output result path, and the target model.
Before any heavy computation, it performs lightweight connection checks against the configured \gls{llms} APIs (OpenAI, Claude, Gemini) to fail fast if credentials are missing or invalid.
The script then enumerates Slack channel subdirectories. 
It iterates over the target users, skips those without contributions (channels where the user is absent) or those with already materialized outputs, and dispatches the collected text to the appropriate chunked API helper. 
Outputs are emitted on a per-user, per-channel basis, accompanied by console-level progress cues.

The domain knowledge acquired for each user across different channels was ultimately averaged with the inference results from other channels to visualize the knowledge level. 
We will now describe these processes in greater detail.


\subsubsection{Model-dependent token parameters.}
Different OpenAI models expose different API parameters for controlling output length.
We therefore explicitly distinguish between models that use a traditional \texttt{max\_tokens} parameter and models that instead use \texttt{max\_completion\_tokens}.
In particular, we treat ``o-series'' models (whose names start with \texttt{o} followed by a digit, e.g., \texttt{o3}, \texttt{o3-mini}, \texttt{o4-mini}) and the GPT-5 family (e.g., \texttt{gpt-5}, \texttt{gpt-5-pro}) as using \texttt{max\_completion\_tokens}, whereas all remaining OpenAI chat models are treated as using \texttt{max\_tokens}.
A helper function automatically checks the normalized model name and chooses the appropriate parameter.
For o-series and GPT-5 models, we also acknowledge that some variants effectively support only the default sampling behaviour, and therefore avoid setting a temperature parameter when it is not meaningful (e.g., for purely deterministic configurations).
For Anthropic Claude, we call the messages endpoint with an explicit upper bound on generated tokens (\texttt{max\_tokens = 4096}), regardless of the context window.
For Google Gemini, we control sampling via a generation configuration that includes the temperature but leaves the maximum number of generated tokens implicit.
At the same time, we explicitly constrain the total prompt size to stay within the model-specific context window.

\begin{figure*}[t!]
  \caption{\gls{llms} prompt template used to estimate domain knowledge from Slack logs during evaluation.}
  \Description{Prompt box that outlines the context, task instructions, and JSON output format provided to the large language models.}
  \label{lst:knowledge_prompt}
  \begin{flushleft}
  \begin{llmPromptBox}{}
  
  \textbf{Prompt Template:}
  \begin{lstlisting}[basicstyle=\ttfamily\small]
  You are an expert in analyzing and estimating a user's domain knowledge based on 
  log data. Focus specifically on the "TARGETUSER" to analyze their knowledge level 
  based on "INPUTDATA". The "TARGETUSER" corresponds to "user" in the "INPUTDATA".
  
  Instructions:
  - Extract domain knowledge by analyzing the "text" fields for the target user.
  - In the output, list proper nouns related to skills, domains, or key terms 
    (e.g., technology, methods, or concepts).
  - For each extracted item, classify the knowledge level:
    - 2 (Known): Strong evidence the user knows this.
    - 1 (Maybe known): Some evidence, moderate confidence.
    - 0 (Unknown): Insufficient evidence of knowledge.
  - For each item, give a brief reason for your classification based on INPUTDATA.

  INPUTDATA:
  {chunk}
  \end{lstlisting}
  
  \textbf{Example Output JSON:}
  \begin{lstlisting}[basicstyle=\ttfamily\small]
  "{target_user_id}": {{
      "text": "Extracted proper noun or verb from the text in INPUTDATA",
      "level": 2 (Known), 1 (Maybe known), or 0 (Unknown),
      "reason": "Brief explanation for why this knowledge level was assigned based
                 on the INPUTDATA"
  }}
  \end{lstlisting}
  \end{llmPromptBox}
  \end{flushleft}
\end{figure*}

\subsubsection{Model-specific context windows.}
To handle long JSON logs, we approximate each model's maximum context length and
derive a per-chunk budget of input tokens.
For OpenAI models, we maintain a table of approximate maximum context sizes (in tokens) for representative models (e.g., \texttt{gpt-4o}, \texttt{gpt-5}, and \texttt{o3}: approximately 128 000 tokens).
If the user does not specify a context window, we infer it by partially matching the normalized model name against this table; otherwise, a default of 4{,}096 tokens is used.
For Claude models, we assume a large context window of 200{,}000 tokens for several recent variants (e.g., \texttt{claude-sonnet-4-5}, and \texttt{claude-haiku-4-5}).
For Gemini, we similarly maintain approximate context windows of 32{,}768 tokens for both \texttt{gemini-2-5-pro} and \texttt{gemini-2-5-flash} models.
If no explicit limit is given, the code infers a maximum from this table based on partial matches of the model name.

\subsubsection{Token counting and chunking strategy.}
To split large JSON logs into model-compatible pieces, we approximate token counts using the \texttt{cl100k\_base} tokenizer.
Although this tokenizer is native to OpenAI models, we also use it as a conservative approximation for Claude and Gemini.
Let $T_{\text{max}}$ denote the assumed maximum context length for a given model, $T_{\text{sys}}$ the number of tokens occupied by the fixed system prompt, and $T_{\text{res}}$ a reserved budget of tokens left free for the model's generated output (default $T_{\text{res}} = 500$).
We then define an effective upper bound on the total tokens we are willing to use for each request by applying a safety factor $s \in (0,1)$ to the context window:
\[
T_{\text{eff}} = \lfloor s \cdot T_{\text{max}} \rfloor.
\]
The per-chunk budget for the user content is then.
\[
T_{\text{chunk}} = T_{\text{eff}} - T_{\text{sys}} - T_{\text{tmpl}} - T_{\text{res}},
\]
where $T_{\text{tmpl}}$ accounts for any fixed tokens in the user message template (e.g., headers such as ``TARGETUSER'' and ``INPUTDATA'').
For OpenAI models, we typically use $s = 0.75$; for Claude, we use a more conservative safety factor (e.g., $s \approx 0.65$) to compensate for tokenizer differences and message overhead; for Gemini, we again use $s = 0.75$.

Given an input log serialized as a JSON string, we encode it into tokens with the approximate tokenizer, compute the total number of tokens $T_{\text{input}}$, and split it into
\[
N_{\text{chunks}} = \left\lceil \frac{T_{\text{input}}}{T_{\text{chunk}}} \right\rceil
\]
contiguous segments.
Each segment is then decoded back into text and passed as the ``INPUTDATA'' for a separate model call.
If the user additionally specifies a hard cap $N_{\text{max}}$ on the number of chunks, we process at most $\min(N_{\text{chunks}}, N_{\text{max}})$ segments and log a warning if the cap is reached.

\subsubsection{Per-provider prompting and execution.}
Across providers, we reuse a common semantic task: estimating a target user's domain knowledge from log data.
\autoref{lst:knowledge_prompt} shows the system prompt used in this study.
We send the prompt together with the Slack communication login token chunk.
The model is instructed to return a single JSON object with entries of the form
\texttt{\{"text": "...", "level": 0|1|2, "reason": "..." \}},
where \texttt{level} encodes whether the knowledge is unknown, maybe known, or clearly known.
For OpenAI, we also enable JSON response formats when supported by the underlying model (e.g., \texttt{gpt-4o} with \texttt{type="json"}, GPT-5 with \texttt{type="json\_object"}). 
In contrast, o-series models are queried without a structured \texttt{response\_format} due to current API constraints.
For Claude and Gemini, we pass the combined prompt (system plus user) in the provider-specific format, while maintaining the token budgeting strategy described above.

\begin{figure*}[t!]
  \centering
  \includegraphics[width=\textwidth]{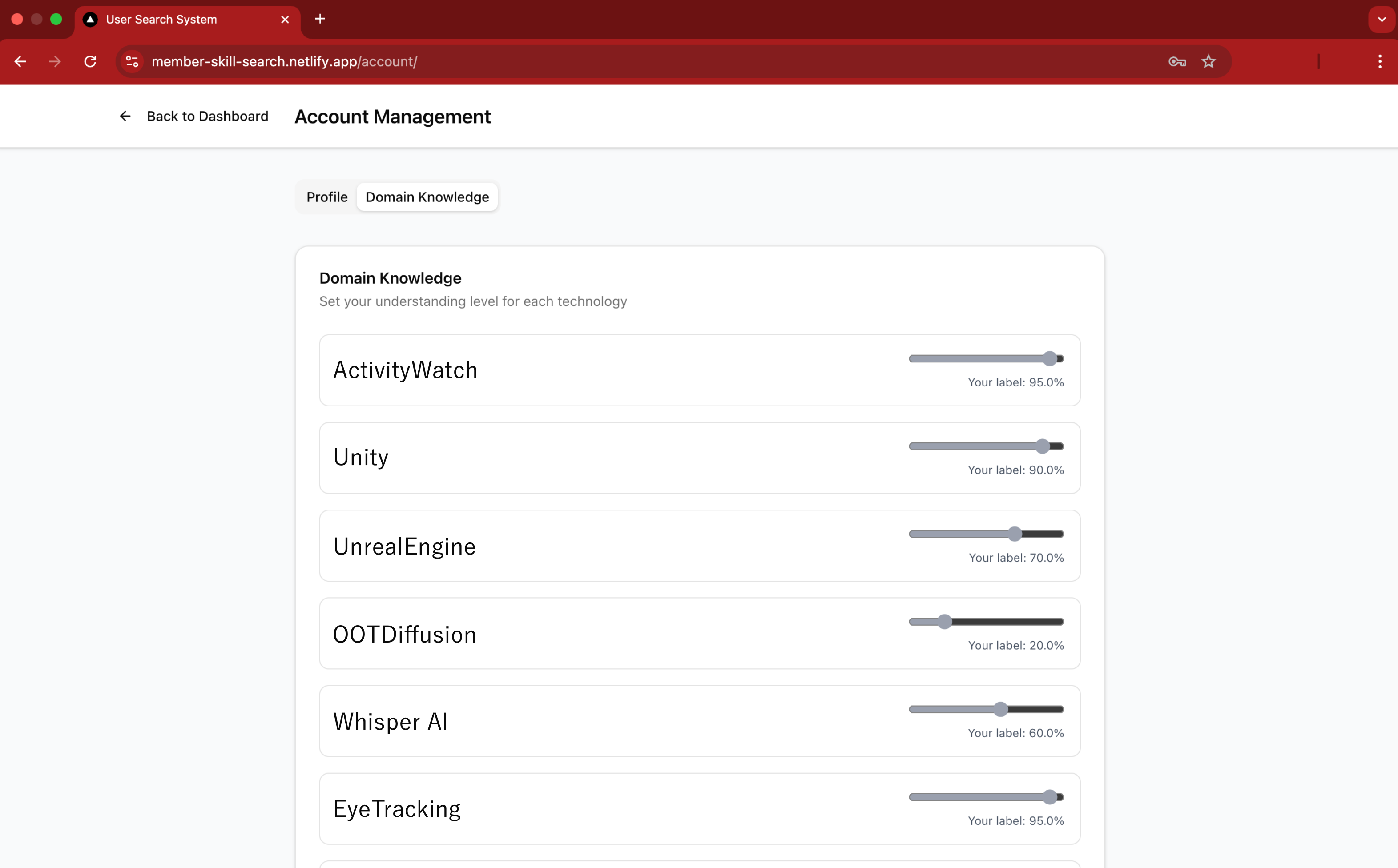}
  \caption{
    Web application user interface (UI) for self-annotation.
    Users were instructed to self-rate their skills, which were displayed on their personal page.
    The domain knowledge (terminology) is shown according to the \gls{llms} extraction.
    }
  \Description{
    Web application user interface (UI) for self-annotation.
    Users were instructed to self-rate their skills, which were displayed on their personal page.
    The domain knowledge (terminology) is shown according to the \gls{llms} extraction.
    }
  \label{fig:ui_self_annotation}
\end{figure*}

\subsection{Performance Evaluation of Domain Knowledge Estimation Through User Self-Annotation}
\label{subsec:performance_metrics}
Self-annotation is the last key point of our study method.
We encourage the Slack users who are active at least once.
The service exposes three GET endpoints that retrieve member profiles and their skills, and a POST endpoint that updates the self-reported skill levels in the Firebase cloud database.
Each endpoint loads group metadata from disk, initializes Firebase credentials lazily, and interacts with Firestore collections to aggregate skills, compute the top-five averages, and merge self-assessments. 
Any failures propagate as HTTP exceptions with explicit status codes, keeping client feedback actionable.
\autoref{fig:ui_self_annotation} shows the user interface (UI) for the self-annotation application.
As shown in the figure, participants were instructed to rate their own skills on a scale of 0 to 100, in increments of 5.

We evaluate the performance of the \gls{llms} through mean absolute error (MAE) and standard deviation of the mean absolute errors (MAE\_STD), root mean square error (RMSE), and median absolute error (Median AE).
Here are the definitions of the performance metrics.

\begin{equation}
\text{MAE} = \frac{1}{n} \sum_{i=1}^{n} |y_i - \hat{y}_i|
\end{equation}

where $y_i$ is the true value, $\hat{y}_i$ is the predicted value, and $n$ is the number of samples.

\begin{equation}
\text{MAE}_{STD} = \sqrt{\frac{1}{n-1} \sum_{i=1}^{n} (|y_i - \hat{y}_i| - \text{MAE})^2}
\end{equation}

where $n$ is the number of samples.

\begin{equation}
\text{RMSE} = \sqrt{\frac{1}{n} \sum_{i=1}^{n} (y_i - \hat{y}_i)^2}
\end{equation}

where $y_i$ is the true value, $\hat{y}_i$ is the predicted value, and $n$ is the number of samples.

\begin{equation}
\text{Median AE} = \text{median}(|y_1 - \hat{y}_1|, |y_2 - \hat{y}_2|, \ldots, |y_n - \hat{y}_n|)
\end{equation}

where $y_i$ is the true value, $\hat{y}_i$ is the predicted value, and $n$ is the number of samples.

\section{Data Collection}
In this section, we will explain the data collection process in detail.
All participants are based in Germany, so we have considered the General Data Protection Regulation (GDPR) in the ethical review process.
We have also obtained the approval from the ethics committee of the \textbf{anonymized for the review process, and this text will be updated once the approval is obtained}.

\subsection{Participants}
\label{subsec:participants}
In this study, the target participants are restricted to the Slack users who are active at least once.
Since some Slack members were former members, we reached 27 participants through follow-up email invitations.
To understand the demographics, we asked all participants to complete the pre-survey. 
The pre-survey results showed that the participants' mean age was 27.96 years (SD = 3.30).
Of these, 22 identified as male (81\%), four as female (15\%), and one preferred not to disclose their gender (4\%).
Regarding nationality, 12 participants were from India and 11 from Japan, with one participant each from Chile, Germany, Iran, and Russia.
Regarding occupation, 18 participants (67\%) were employed, and 9 (33\%) were students.

\begin{figure*}[t!]
  \centering
  \includegraphics[width=\textwidth]{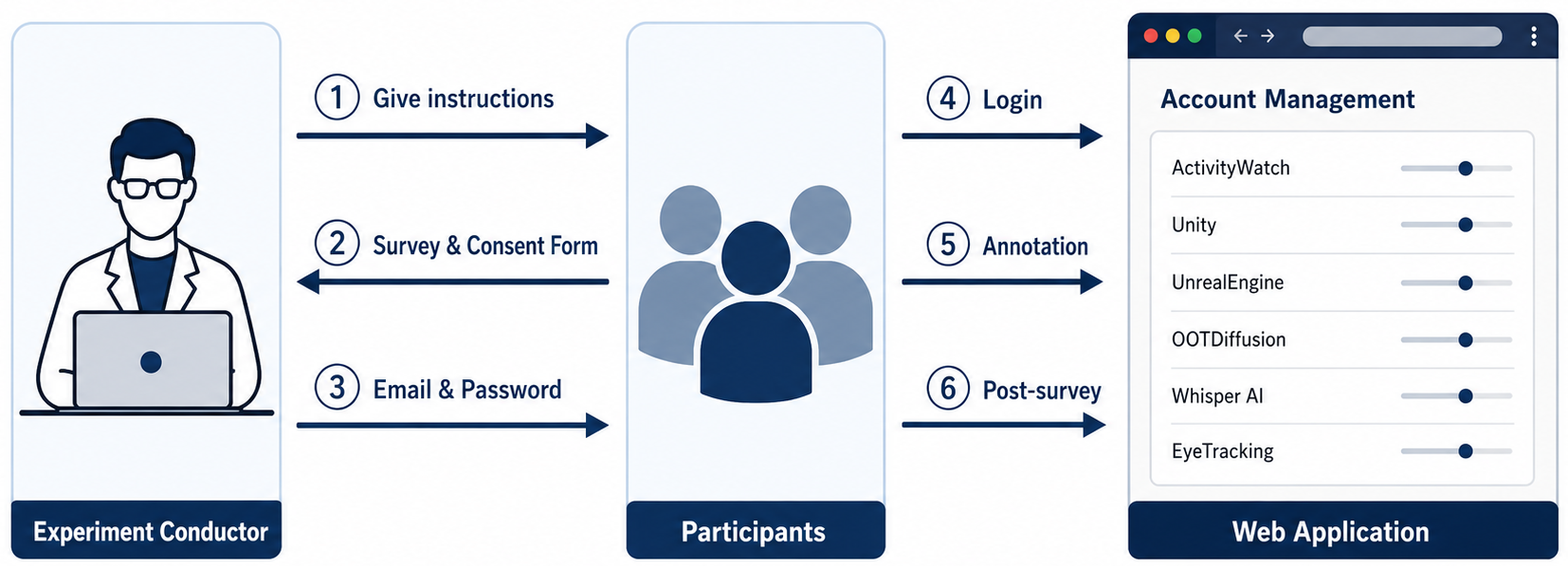}
  \caption{
    Experiment procedure.
    Participants first listen to the experiment instructions from the experiment conductor.
    The user then confirms and gives the signed consent form and the online survey. 
    Once those are completed, the experiment conductor shares the login information for each participant.
    Then, participants login to the web application and self-rate their skills.
    Finally, participants complete the post-survey.
    }
  \Description{
    Experiment procedure.
    Participants first listen to the experiment instructions from the experiment conductor.
    User then confirm and give the signed consent form and the online survey. 
    Once those are completed, experiment conductor share the login information for each participant.
    Then, participants login to the web application and self-rate their skills.
    Finally, participants complete the post-survey.
    }
  \label{fig:experiment_procedure}
\end{figure*}

\subsection{Experiment Procedure}
\autoref{fig:experiment_procedure} shows the experiment procedure.
Participants were first asked to receive instructions from the experiment conductor.
This instruction includes the provision that participants may opt out at any time.
Once the participants agree to the experiment, participants fill out the consent form and the survey~\footnote{Google Form: \url{https://forms.gle/DjXDPAxGKzDyXFeQ6}}.
The ethics committee evaluates the consent form, and the survey is used only for demographic information (\autoref{subsec:participants}).

Once the preparatory phase was complete, participants were asked to login to the web application~\footnote{Web Application: \url{https://member-skill-search.netlify.app/}} on their own laptop.
The experiment conductor primarily created each user's account (email and password) and distributed it to each participant.
Email is the one used for Slack login, so this is the approach to connect each participant's domain knowledge.
Once participants logged in to the web application, they were asked to self-rate their skills.
The domain knowledge (terminology) is shown according to the \gls{llms} extraction.
To avoid biasing the \gls{llms} estimation, the application did not display the skill level estimated by the \gls{llms}.
After the self-annotation task was complete, participants were asked to complete the post-survey~\footnote{Google Form: \url{https://forms.gle/vcWx2VrCLfqRej5z9}}.
The post-survey collects participants' feedback on their general thoughts about this study.

\begin{figure*}[t!]
  \centering
  \includegraphics[width=\textwidth]{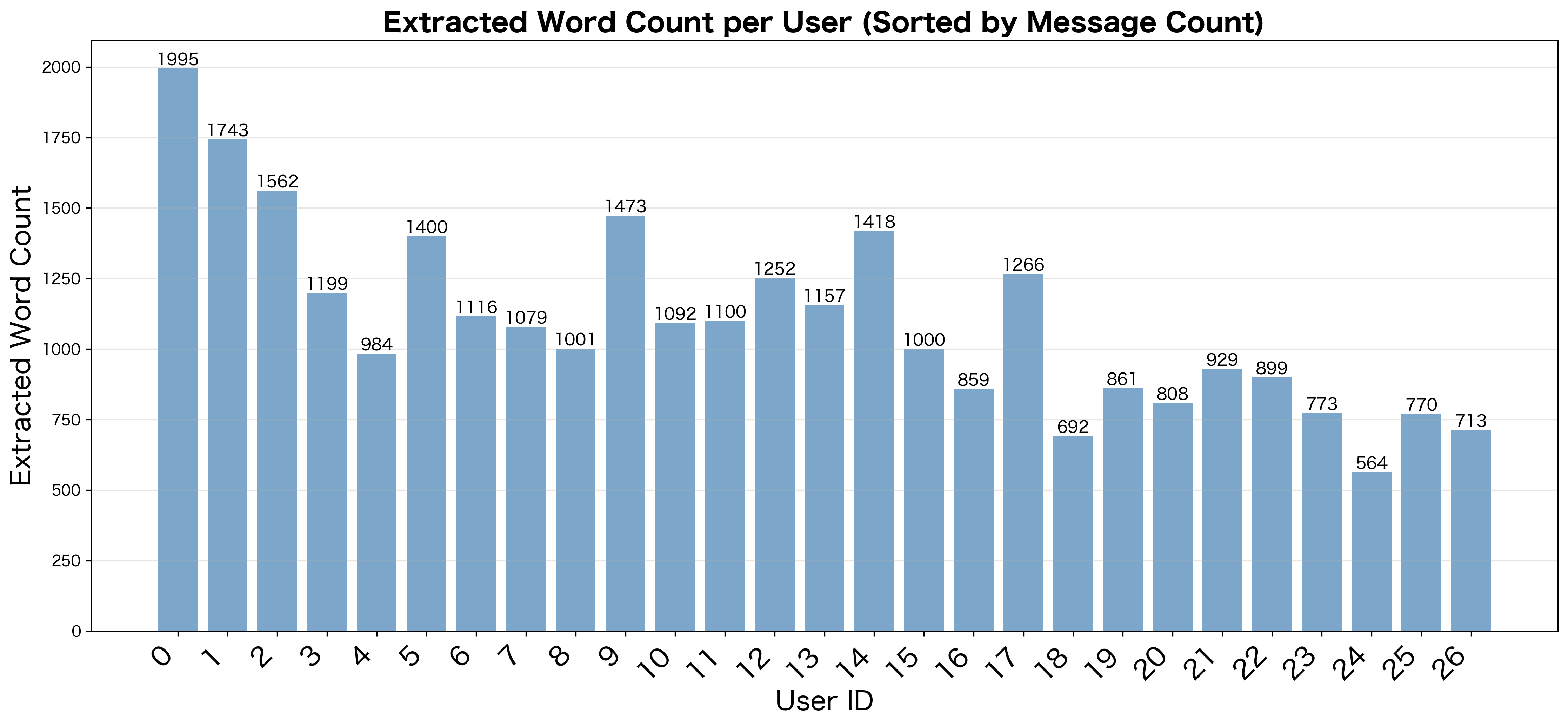}
  \caption{
    The number of words extracted from the communication logs for each participant.
    The bar chart shows the number of words extracted from each participant's communication logs.
    The number of words is the sum of the number of words in the message content and the number of words in the message metadata.
    }
  \Description{
    The number of words extracted from the communication logs for each participant.
    The number of words is the sum of the number of words in the message content and the number of words in the message metadata.
    }
  \label{fig:extracted_word_count_per_user}
\end{figure*}

\section{Result and Discussion}
This section presents the study's findings, organized by research questions.
\autoref{fig:extracted_word_count_per_user} shows the amount of domain knowledge extracted from the communication logs for each participant, sorted by the number of user messages.
In general, participants who sent more messages had more domain knowledge extracted from their logs.
However, this relationship is not strictly linear, as it also depends on the number of messages across the channels each participant accesses.

\subsection{Performance Evaluation of LLMs}
We first address RQ1: ``How precise can \gls{llms} estimate the human domain knowledge?''.
In this analysis, we treat each participant's self-annotated skill level on the 0--100 scale as ground truth and compare it with the corresponding estimates produced by each model. 
Following the evaluation protocol described in \autoref{subsec:performance_metrics}, we compute the mean absolute error (MAE), its standard deviation (MAE\_STD), the root mean square error (RMSE), and the median absolute error (Median AE) over all skills and all participants.
These metrics quantify different aspects of the discrepancy between \gls{llms}-estimated domain knowledge and self-reported domain knowledge: MAE and Median AE capture typical deviations on the original rating scale, RMSE emphasizes larger errors, and MAE\_STD reflects how consistently a model performs across individual skill items.

\autoref{tab:global_metrics_micro} summarizes the overall performance of all evaluated models. 
Across the seven \gls{llms}, we observe that all models exhibit a non-trivial ability to approximate participants' domain knowledge, but the estimates remain far from perfect. 
Gemini 2.5 Flash achieves the best performance, with an MAE of $21.13 \pm 19.14$, an RMSE of $28.48$, and a Median AE of $15.00$. 
On our 0--100 rating scale, this means that, on average, the model's predictions differ from users' self-ratings by about 21 points, and half of the predictions fall within 15 points of the self-annotated values. 
Gemini 2.5 Pro shows slightly larger errors (MAE $= 22.68$, RMSE $= 31.23$, Median AE $= 16.50$), but remains close to Gemini 2.5 Flash, suggesting that both Gemini models capture broadly similar patterns in the communication logs. 

The Claude models occupy the middle of the performance spectrum: Claude Haiku 4.5 yields an MAE of $26.74$ and Claude Sonnet 4.5 an MAE of $27.95$, with RMSE and Median AE values that are likewise intermediate between the Gemini and GPT families. 
In contrast, the GPT models show the largest discrepancies, with MAE values around $32--33$ and RMSE values above $41$, indicating noticeably less accurate alignment with participants' self-annotated knowledge levels. 
The relatively large MAE\_STD values across all models (approximately $19--27$) further highlight substantial variability at the level of individual skills and participants, implying that some domains are consistently easier for the models to estimate than others.

These findings also answer RQ2: ``Which \gls{llms} model provides the most accurate knowledge estimation?''.
The results reveal a clear performance ranking in our setting: Gemini models perform best overall, followed by Claude models, and finally GPT models.
While the absolute error levels indicate that reliable, fine-grained estimation of human domain knowledge remains challenging, the consistent advantage of the Gemini models suggests that architectural or training differences between \gls{llms} families can have a meaningful impact on this type of estimation task.

\begin{table}[t!]
\centering
\caption{Performance Comparison of the \gls{llms} for Domain Knowledge Estimation.}
\label{tab:global_metrics_micro}
  \begin{tabular}{lcccc}
\toprule
    model & mae ($\downarrow$) & mae\_std ($\downarrow$) & rmse ($\downarrow$) & median\_ae ($\downarrow$) \\
\midrule
    gpt-4o & 33.38 & 25.24 & 41.81 & 30.00 \\
    gpt-o3 & 32.71 & 26.55 & 42.09 & 28.50 \\
    gpt-5 & 32.70 & 26.83 & 42.26 & 28.50 \\
    claude-sonnet-4-5 & 27.95 & 22.27 & 35.70 & 25.00 \\
    claude-haiku-4-5 & 26.74 & 22.41 & 34.85 & 25.00 \\
    gemini-2.5-pro & 22.68 & 21.52 & 31.23 & 16.50 \\
    \textbf{gemini-2.5-flash} & \textbf{21.13} & \textbf{19.14} & \textbf{28.48} & \textbf{15.00} \\
\bottomrule
\end{tabular}
\end{table}

\begin{figure*}[t!]
  \centering
  \includegraphics[width=\textwidth]{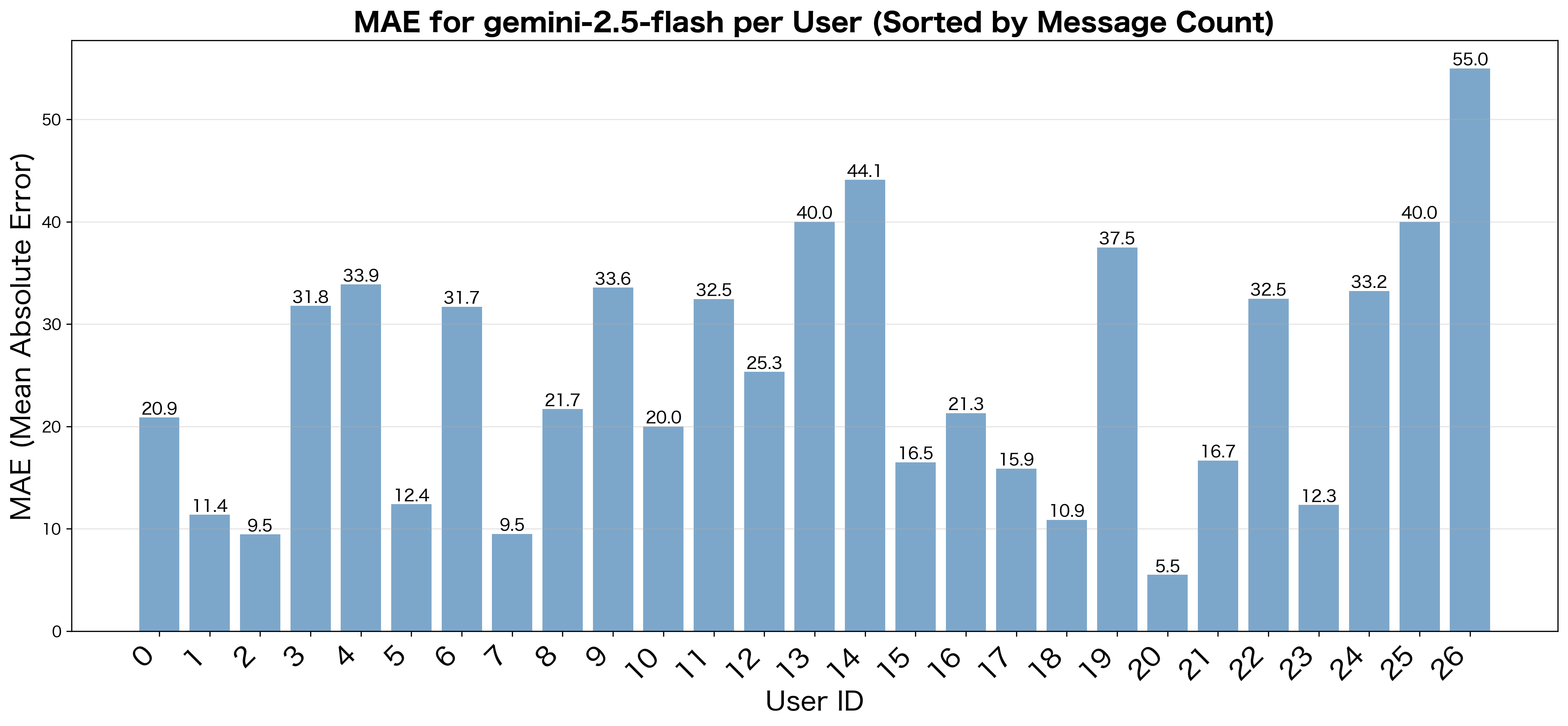}
  \caption{
    Performance comparison of \gls{llms} domain knowledge estimation.
    The bar chart shows the mean absolute error ($\downarrow$) calculated from the errors obtained through self-annotation by LLMs and humans for all participants' data.
    As shown in the result, Gemini 2.5 flash performed the lowest mean absolute error against self-annotated value.
    The most lowest performance was GPT 4o.
    }
  \Description{
    Performance comparison of \gls{llms} domain knowledge estimation.
    The bar chart shows the mean absolute error calculated from the errors obtained through self-annotation by LLMs and humans for all participants' data.
    As shown in the result, Gemini 2.5 flash performed the lowest mean absolute error against self-annotated value.
    The most lowest performance was GPT 4o.
    }
  \label{fig:gemini_perfomance_sort_with_message_count}
\end{figure*}

\subsection{Performance Comparison Among Individuals}
We now address RQ3: ``How does the amount of communication logs impact the accuracy of \gls{llms} domain knowledge estimation?'' 
As described in \autoref{subsec:log_statics}, the number of Slack messages per participant varies substantially, from only a handful of posts to more than ten thousand messages, providing a natural testbed for analyzing how data volume relates to estimation performance. 
To investigate this relationship, we focus on Gemini 2.5 Flash, the best-performing model in the aggregate evaluation, and compute the MAE between its estimated skill scores and each participant's self-annotated scores. 
\autoref{fig:gemini_perfomance_sort_with_message_count} visualizes these per-user MAE values, with participants ordered by their total number of user messages.

Overall, the figure reveals considerable variability in MAE across individuals, largely independent of their message volume. 
Participants with very few messages tend to exhibit relatively large errors, suggesting that when the available communication history is limited, \gls{llms} lack sufficient evidence to infer domain knowledge reliably. 
Beyond this low-activity regime, however, the MAE fluctuates within a similar range even as the number of messages increases by an order of magnitude or more. 
In other words, we do not observe a systematic improvement in estimation accuracy for high-activity users; some participants with many messages still show MAE values comparable to, or even higher than, those of participants with moderate message counts.

This pattern indicates that, under our current zero-shot setup, simply accumulating more messages is not sufficient to guarantee more accurate domain-knowledge estimation. 
One plausible explanation is that not all messages are equally informative: a substantial fraction of Slack communication consists of brief acknowledgements, social talk, or coordination messages that convey little about a user's technical expertise. 
In addition, the Slack logs capture only part of each participant's work and study activities, whereas the self-annotated skill scores may include knowledge rarely verbalized in chat.
Finally, because we deliberately did not fine-tune or condition the \gls{llms} on any task-specific examples, the models had to infer expertise patterns purely from their general prior knowledge and the raw logs. 
Taken together, these factors likely limit the benefits of additional messages and suggest that future work should explore more targeted prompting, lightweight personalization, or task-specific adaptation to exploit the available communication data better.

\section{Limitation and Future Work}

\subsection{Data Security and Privacy}
A central concern in this study is the security and privacy of the communication data used for analysis. 
As described in \autoref{subsec:log_statics}, our dataset consists of Slack communication logs collected over 2{,}744 days, comprising 27{,}188 messages from 43 users across 94 channels within a single group. 
Because these logs may contain both corporate confidential information and personal data, they must be treated as highly sensitive.
All participants were based in Germany, and the data collection and analysis procedures were designed in accordance with the General Data Protection Regulation (GDPR) and an institutional ethics review process.

In the present work, these logs are used solely for research purposes to investigate whether \gls{llms} can estimate individual domain knowledge from past communication.
Only users who could be contacted and who later provided informed consent (27 out of the 43 Slack accounts) are included as participants in our evaluation.
Their self-annotated skill scores are linked to the \gls{llms}-estimated scores, and all analyses are conducted using pseudonymized user identifiers (UIDs) rather than real names.
User information is represented by IDs such as UID 0, and message examples shown in the paper are partially anonymized.
These design choices aim to minimize the exposure of identifiable or sensitive content while still allowing us to study the behavior of different models on realistic organizational data.

At the same time, our experimental setup relies on cloud-based APIs for the evaluated \gls{llms} families (OpenAI, Claude, and Gemini), meaning that segments of the Slack logs are sent to external providers during inference.
Under our current research protocol and consent procedure, this is acceptable for assessing the feasibility and relative performance of different \gls{llms} for domain-knowledge estimation. 
However, many organizations operate under stricter data-governance policies or data-residency requirements, where sending internal communication logs to third-party cloud services would be unacceptable. 
Consequently, the present study should be interpreted as a proof-of-concept demonstration rather than a ready-to-deploy solution for highly sensitive environments.

As an important direction for future work, we plan to explore deployment strategies that keep all communication data within the organization's own security boundary. 
One promising approach is to employ local or self-hosted \gls{llms}, for example, models deployed on on-premises servers or in tenant-isolated environments, so that raw communication logs never leave corporate infrastructure.
Complementary to this, we aim to investigate privacy-preserving preprocessing techniques, such as stronger anonymization or message aggregation, as well as the use of intermediate user representations that can be processed by \gls{llms} without exposing the original message content.
Developing such privacy-aware variants of our method will be essential for making \gls{llms}-based domain-knowledge estimation practically applicable in real-world organizations that must balance knowledge sharing with rigorous security and privacy constraints.

\subsection{Towards Practical Application}
In this study, we took a first step toward practical use by investigating whether individual domain knowledge can be estimated from everyday organizational communication, specifically Slack logs, using \gls{llms}.
Our experimental setup focused on how well models can infer users' expertise from organically occurring messages, without requiring additional structured inputs or explicit self-descriptions. 
To make the process manageable for participants, we limited the target of estimation to word-level skill terms (e.g., technology names or key concepts). We asked users to provide self-assessed familiarity scores on a 0--100 scale via a dedicated web application. 
This design enabled a systematic, quantitative evaluation across models while keeping the annotation burden relatively low.

At the same time, our findings highlight several gaps that need to be addressed before such an approach can be deployed in real-world workflows.
Domain knowledge is inherently multifaceted and context-dependent; it involves not only familiarity with isolated concepts but also the depth of understanding, the ability to combine skills across domains, and experience with specific tools, projects, or roles.
Representing expertise purely as a flat list of word-level skills therefore overlooks meaningful relationships among skills, differences in seniority or specialization, and the temporal evolution of knowledge. 
Moreover, organizational communication channels may be noisy and incomplete, with substantial variation in how individuals express their expertise, leading to uneven estimation performance across users and domains.

As future work, we plan to design a more general extraction pipeline that captures richer structures of competence while preserving interpretability and ease of use. 
For example, we envision methods that derive skill clusters, topic hierarchies, or role-specific profiles from communication logs, combined with other data sources such as project repositories or internal documentation. 
Such representations support more nuanced reasoning about who knows what, including the identification of complementary expertise and emerging specialists. 
Building on these representations, our long-term goal is to develop a practical mechanism in which, when an employee submits a question or task, the backend system automatically leverages inferred domain-knowledge profiles to identify and recommend suitable experts within the organization. 
In this way, \gls{llms}-based estimation of domain knowledge could serve as a foundation for expert-finding tools and knowledge-support systems that augment existing communication channels, analogous to how current \gls{llms} prompts route user queries to appropriate capabilities.

\section{Conclusion}
In this study, we investigated whether LLMs can estimate domain knowledge from organizational chat logs. Using 27,188 Slack messages from 43 users over 2,744 days, we built a pipeline that extracts skill terms via LLMs and aggregates them into per-user knowledge profiles, which 27 participants then self-rated. Our results show that contemporary LLMs can approximate human domain knowledge meaningfully but remain unreliable for fine-grained profiling. Across seven models, mean absolute errors ranged from 21 to 33 points on a 0–100 scale. Gemini 2.5 Flash achieved the best performance (MAE 21.13), followed by Gemini 2.5 Pro, with Claude models in the middle and GPT models trailing. Per-user analysis revealed substantial individual variation and only weak dependence on message volume: sparse histories hurt performance, but beyond a minimal threshold, additional messages did not systematically reduce errors. These findings highlight both the promise and current limitations of using LLMs to infer ``who knows what'' from communication logs, providing a foundation for future research on AI-supported organizational knowledge sharing.


\bibliographystyle{ACM-Reference-Format}
\bibliography{main}
\end{document}